\documentclass{article} % For LaTeX2e
\usepackage{misc/iclr2020_conference}
\iclrfinalcopy % Uncomment for camera-ready version

\usepackage{enumitem}
\usepackage{natbib}
\setcitestyle{authoryear,open={[},close={]}}
\usepackage{lipsum}
\usepackage{xcolor}
\definecolor{darkblue}{rgb}{0.0, 0.0, 0.55} % https://latexcolor.com/
\definecolor{darkcerulean}{rgb}{0.03, 0.27, 0.49}
\definecolor{darkcandyapplered}{rgb}{0.64, 0.0, 0.0}
\definecolor{darklavender}{rgb}{0.45, 0.31, 0.59}
\definecolor{darkmagenta}{rgb}{0.55, 0.0, 0.55}

\usepackage[utf8]{inputenc}
\usepackage[english]{babel}
\usepackage[T1]{fontenc}
\usepackage{microtype}      % microtypography
\usepackage{times}

\usepackage{graphicx,adjustbox,subcaption,wrapfig}
\usepackage{multirow,multicol,booktabs}
\usepackage{url}
\usepackage[colorlinks=true,
            linkcolor=darkblue,
            citecolor=darkblue,
            filecolor=darkblue,
            urlcolor= darkmagenta,]{hyperref}
\usepackage{bookmark} % !!! loaded after hyperref to avoid conflicts

\newcommand{\etal}{\emph{et al. }}

\usepackage[ruled,noline,linesnumbered]{algorithm2e}
\IncMargin{1em}
\usepackage{listings}
\definecolor{codegreen}{rgb}{0,0.6,0}
\definecolor{codegray}{rgb}{0.5,0.5,0.5}
\definecolor{codepurple}{rgb}{0.58,0,0.82}
\definecolor{backcolour}{rgb}{0.95,0.95,0.92}
\lstdefinestyle{mystyle}{
    backgroundcolor=\color{backcolour},   
    commentstyle=\color{codegreen},
    keywordstyle=\color{magenta},
    numberstyle=\tiny\color{codegray},
    stringstyle=\color{codepurple},
    basicstyle=\ttfamily\footnotesize,
    breakatwhitespace=false,         
    breaklines=true,                 
    captionpos=b,                    
    keepspaces=true,                 
    showspaces=false,                
    showstringspaces=false,
    showtabs=false,                  
    tabsize=2
    }
\usepackage{amsmath,amsfonts, mathtools, bm, bbm, nicefrac}
\usepackage[thinc]{esdiff}
\newcommand{\dd}{\mathrm{d}}

\newcommand{\E}{\mathbb{E}}

\def\rvz{{\mathbf{z}}}

\def\vzero{{\bm{0}}}

\def\vc{{\bm{c}}}

\def\vv{{\bm{v}}}

\def\vz{{\bm{z}}}

\def\mI{{\bm{I}}}

\DeclareMathAlphabet{\mathsfit}{\encodingdefault}{\sfdefault}{m}{sl}
\SetMathAlphabet{\mathsfit}{bold}{\encodingdefault}{\sfdefault}{bx}{n}

\title{PeRFlow: Piecewise Rectified Flow as Universal Plug-and-Play Accelerator}
\author{
Hanshu Yan\textsuperscript{*}, 
~Xingchao Liu\textsuperscript{+}, 
~Jiachun Pan\textsuperscript{\#}, 
~Jun Hao Liew\textsuperscript{*}, 
~Qiang Liu\textsuperscript{+}, 
~Jiashi Feng\textsuperscript{*}\\
\textsuperscript{\textbf{*}}ByteDance, 
~\textsuperscript{\textbf{+}}Univeristy of Texas at Austin, 
~\textsuperscript{\textbf{\#}}National University of Singapore\\
~{\small hanshu.yan@outlook.com}
}

\begin{document}
\maketitle

\begin{abstract}
We present Piecewise Rectified Flow (PeRFlow), a flow-based method for accelerating diffusion models. PeRFlow divides the sampling process of generative flows into several time windows and straightens the trajectories in each interval via the reflow operation, thereby approaching piecewise linear flows. PeRFlow achieves superior performance in a few-step generation. Moreover, through dedicated parameterizations, the PeRFlow models inherit knowledge from the pretrained diffusion models. Thus, the training converges fast and the obtained models show advantageous transfer ability, serving as universal plug-and-play accelerators that are compatible with various workflows based on the pre-trained diffusion models. Codes for training and inference are publicly released. \footnote{\url{https://github.com/magic-research/piecewise-rectified-flow}}.
\end{abstract}

\section{Introduction}
Diffusion models have exhibited impressive generation performances across different modalities, such as image~\citep{rombach_high-resolution_2022, ho2022cascaded, song_score-based_2021, balaji2022ediff}, video~\citep{ho2022video, zhou_magicvideo_2023, wang2024magicvideo, liew2023magicedit, xu2023magicanimate}, and audio~\citep{kong2020diffwave}. Diffusion models generate samples by reversing pre-defined complicated diffusion processes, thus requiring many inference steps to synthesize high-quality results. Such expensive computational cost hinders their deployment \citep{li2024snapfusion, song2023consistency, pan2023adjointdpm} in real-world applications.  

%The sampling processes of diffusion models can be formulated as 
%solving the corresponding probability-flow ordinary differential equations (PF-ODEs) \citep{song_score-based_2021}. 
Diffusion models can be efficiently sampled by solving the corresponding probability flow ordinary differential equations (PF-ODEs)~\citep{song_score-based_2021, song_denoising_2022}.
Researchers have designed many advanced samplers, such as DDIM \citep{song_denoising_2022}, DPM-solver \citep{lu_dpm-solver_2022}, and DEIS \citep{zhang2022fast}, to accelerate generation, inspired by the semi-linear structure and adaptive solvers in ODEs. However, these samplers still require tens of inference steps to generate satisfying results. Researchers have also explored distilling pretrained diffusion models into few-step generative models \citep{salimans_progressive_2022, meng2023distillation, gu2023boot, yin2023onestep, nguyen2024swiftbrush, berthelot2023tract}, which have succeeded in synthesizing images within 8 inference steps. 
Progressive Distillation \citep{salimans_progressive_2022} separates the whole sampling process into multiple segments and learns the mapping from starting points to endpoints for each segment.
Distribution Matching Distillation~\citep{yin2023onestep} and SwiftBrush~\citep{nguyen2024swiftbrush} use the score distillation loss to align the distributions of teacher and one-step student generators.
UFOGen~\citep{xu2023ufogen}, SDXL-Turbo~\citep{sauer2023adversarial} and SDXL-Lightning~\citep{lin2024sdxllightning} resort to adversarial training for learning few-step/one-step image generators. They initialize the students from pretrained diffusion models and use adversarial and/or MSE losses to align the student model's generation with the pretrained ones. 
These methods suffer from the difficult tuning of the adversarial training procedure and the mode collapse issue.
Latent Consistency Model (LCM) ~\citep{luo2023latent, luo2023lcmlora} adopts consistency distillation~\citep{song2023consistency} to train a generator that directly maps noises to the terminal images. 
LCM only utilizes supervised distillation where the training procedure will be more stable and easier in comparison to adversarial training.
However, the generated images have fewer details compared with SDXL-Lighting.

Unlike the existing methods above, which mainly learn the mappings from noises to images, we aim to simplify the flow trajectories and preserve the continuous flow trajectories of the original pretrained diffusion models. Specifically, we attempt to straighten the trajectories of the original PF-ODEs via a piecewise reflow operation. 
Previously, InstaFlow~\citep{liu2023instaflow} leverages the rectified flow framework\citep{liu2022flow, liu2022rectified} to learn the transformation from initial random noise to images.
It bridges the two distributions with linear interpolation and trains the model by matching the interpolation.
With the reflow operation, it may be able to learn straight-line flows for one-step generation via pure supervised learning. 
%, 
%where the reflow operation aims to build a linear bridge between two distributions. 
% InstaFlow samples a large set of noise-text pairs and then calls the pretrained diffusion models to synthesize images for each pair. The sampled noises and generated images form two distributions. The linear flow is trained via a velocity-matching MSE loss. 
InstaFlow provides a simple pipeline for accelerating pretrained diffusion models, however, it suffers from poor sampling quality which can be attributed to synthetic data generation. The reflow operation requires generating data from the pretrained diffusion models with ODE solvers (e.g., DDIM or DPM-Solver~\citep{lu_dpm-solver_2022, lu2023dpmsolver}) to construct a training dataset. Synthesizing training data brings two problems: (1) constructing and storing the dataset requires excessive time and space, which limits its training efficiency; (2) synthetic data has a noticeable gap with real training data
in quality due to the numerical error of solving ODEs. Thus, the performance of the learned straighter flow is bounded. 

To address the problems, \textit{we propose piecewise rectified flow (PeRFlow), which divides the flow trajectories into several time windows and conducts reflow in each window}. By solving the ODEs in the shortened time interval, PeRFlow avoids simulating the entire ODE trajectory for preparing the training data. This significantly reduces the target synthesis time, enabling the simulation to be performed in real time along with the training procedure. Besides, PeRFlow samples the starting noises by adding random noises to clean images according to the marginal distributions, and 
solves the endpoints of a shorter time interval, which has a lower numerical error than integrating the entire trajectories.
Through such a divide-and-conquer strategy, PeRFlow can straighten the sampling trajectories with large-scale real training data. 
Besides the training framework, \textit{we also design a dedicated parameterization method for PeRFlow to inherit sufficient knowledge from the pretrained diffusion models}. 
Diffusion models are usually trained with $\epsilon$-prediction, but flow-based generative models generate data by following the velocity field. We derive the correspondence between $\epsilon$-prediction and the velocity field of flow, thus narrowing the gap between the pretrained diffusion models and the student PeRFlow model.
Consequently, PeRFlow acceleration converges fast and the resultant model can synthesize highly-detailed images within very few steps.
PeRFlow does not require unstable adversarial training or a complete modification of the training paradigm. It is a lightweight acceleration framework and can be easily applied to training unconditional/conditional generative models of different data modalities.

We conducted extensive experiments to verify the effectiveness of PeRFlow on accelerating pretrained diffusion models, including Stable Diffusion (SD) 1.5, SD 2.1, SDXL \citep{podell2023sdxl}, and AnimateDiff \citep{guo2023animatediff}. 
PeRFlow-accelerated models can generate high-quality results within four steps.
Moreover, we find that the variation of the weights, $\Delta W = \theta-\phi$, between the trained student model $\theta$ and the pretrained diffusion model $\phi$, can serve as universal accelerators of almost all workflows that are only trained on the pretrained diffusion models. These workflows include customized SD models, ControlNets, and multiview 3D generation.
%Here, $\Delta W$ can also be parameterized by LORA \citep{hu2021lora} adaptors. 
We compared PeRFlow with state-of-the-art acceleration methods. PeRFlow shows advantages in terms of FID values, visual quality, and generation diversity. 

In summary, PeRFlow has the following favorable features: 1) it is simple and flexible for accelerating various diffusion pipelines with fast convergence; 2) The accelerated generators support fast generation; 3) The obtained $\Delta W$ shows superior plug-and-play compatibility with the workflows of the pretrained models.

\section{Methodology}

\subsection{Rectified Flow and Reflow}
% A diffusion process, which gradually transforms data distribution $q(\cdot)$ into random noise $\mathcal{N}(\vzero, \mI)$, is pre-defined by a noise schedule $\sigma(t)$. Specifically, the intermediate state $\vz_t = \sqrt{1-\sigma^2(t)} \vz_0 + \sigma(t) \epsilon$, where $\vz_0 \sim q(\cdot)$ and $\epsilon \sim \mathcal{N}(\vzero, \mI)$. The noise schedule $\sigma(t)$ is a monotonically increasing function with $\sigma(0)=0$ and $\sigma(1)=1$. The diffusion generative model aims to learn the reversed process that transports random noise into the data distribution. Then, one can sample new data by solving the corresponding probability-flow ODE. The network $f_{\phi}$ is trained by minimizing noise- or score-matching losses. Due to the complexity of the sampling trajectories, solving probability-flow ODEs usually requires tens of inference steps, which results in extremely high computational costs. 

Flow-based generative models aim to learn a velocity field $\vv_{\theta}(\vz_t,t)$ that transports random noise $\vz_1 \sim \pi_1$ sampled from a noise distribution into certain data distribution $\vz_0 \sim \pi_0$. Then, one can generate samples by solving \eqref{eq:flow-ode} from $t=1$ to $0$:
\begin{align}
    \dd \vz_t = \vv_{\theta}(\vz_t,t) \dd t, \quad \vz_1 \sim \pi_1.
    \label{eq:flow-ode}
\end{align}
% that transports random noise $\vz_1 \sim \pi_1$ into certain data distribution $\vz_0 \sim \pi_0$. 
Recently, simulation-free learning of flow-based models has become prevalent~\citep{liu2022flow, liu2022rectified, lipman2022flow, albergo2023stochastic}. 
A representative method is Rectified flow \citep{liu2022flow, liu2022rectified, lipman2022flow}, which adopts linear interpolation between the noise distribution $\rvz_1$ and the data distribution $\rvz_0$. It trains a neural network $\vv_{\theta}$ to approximate the velocity field via the conditional flow matching loss. The corresponding optimization procedure is termed reflow~\citep{liu2022flow, liu2022rectified},
\begin{align}
    \min_{\theta} \E_{\vz_1 \sim \pi_1, \vz_0 \sim \pi_0} \left[ \int_0^1 \|(\vz_1-\vz_0) - v_{\theta}(\vz_t, t)\|^2 \dd t \right], \quad \text{with}\quad \vz_t = (1-t) \vz_0 + t \vz_1.
    \label{eq:reflow}
\end{align}

% A favorable property of the reflow operation is that, by using the optimal $v_{\theta^*}$ learned in~\eqref{eq:reflow} as teacher, one can repeat the reflow operation to learn a new flow-based generative model which is straighter and faster in sampling.
% \begin{equation}
% \begin{aligned}
%     \min_{\theta'} \E_{\vz_1 \sim \pi_1} &\left[ \int_0^1 \|(\vz_1-\vz_0) - v_{\theta'}(\vz_t, t)\|^2 \dd t \right], \\ \text{with} \quad \vz_0 = \vz_1 + \int_1^0 &v_{\theta^*}(\vz_t, t) \dd t \quad \text{and} \quad \vz_t = (1-t) \vz_0 + t \vz_1.
%     \label{eq:reflow-reflow}
% \end{aligned}
% \end{equation}
 
% Pretrained diffusion models can be transformed to probability flow ODEs, hence they can serve as $v_{\theta^*}$ for learning the straighter flow.
InstaFlow~\citep{liu2023instaflow} proposed to accelerate pretrained diffusion-based text-to-image models via reflow. Given a pretrained diffusion model $f_{\phi}$, one can generate new data by solving the corresponding probability flow ODE. We denote $\Phi(\vz_t, t, s)$ as the ODE solver, such as the DPM-Solver~\citep{lu2022dpm}. For simplicity, our notation drops the parameters in the ODE solvers. 
By simulating with $\vz_0 = \Phi(\vz_1, 1, 0)$, where $\vz_1$ is sampled from the random Gaussian distribution $\pi_1$,
it synthesizes a dataset of \texttt{(text, noise, image)} triplets for reflow. 
Since it usually takes tens of inference steps to generate high-quality data with $\Phi(\vz_1, 1, 0)$, InstaFlow is expensive to scale up. Moreover, since InstaFlow is trained with generated images, it lacks the supervision of real data and thus compromises the resulting generation quality. In the following subsections, we target solving these problems.

\subsection{Piecewise Rectified Flow}
\label{sec:perflow}
We present Piecewise Rectified Flow (PeRFlow), aiming at training a piecewise linear flow to approximate the sampling process of a pretrained diffusion model. PeRFlow sticks to the idea of trajectory straightening. It further allows using high-quality real training data and one-the-fly optimization. PeRFlow is easier to scale up and succeeds in accelerating large-scale diffusion models, including the Stable Diffusion family.

A pretrained diffusion model $f_{\phi}$ corresponds to a probability flow ODE defined by a noise schedule $\sigma(t)$. In the Stable Diffusion family, the forward diffusion process follows $\vz_t = \sqrt{1-\sigma^2(t)} \vz_0 + \sigma(t) \epsilon$, where $\vz_0$ and $\epsilon$ are sampled from the data distribution and random Gaussian respectively. The sampling trajectories are usually complicated curves. Even for an advanced ODE solver $\Phi(\vz_t, t, s)$, it still requires many steps to generate an artifact-free image. We accelerate the pretrained model by applying a divide-and-conquer strategy, that is, we divide the ODE trajectories into multiple time windows and straighten the trajectories in each time window via the reflow operation.

\begin{figure}[!t]
    \centering
    \includegraphics[width=0.9\textwidth]{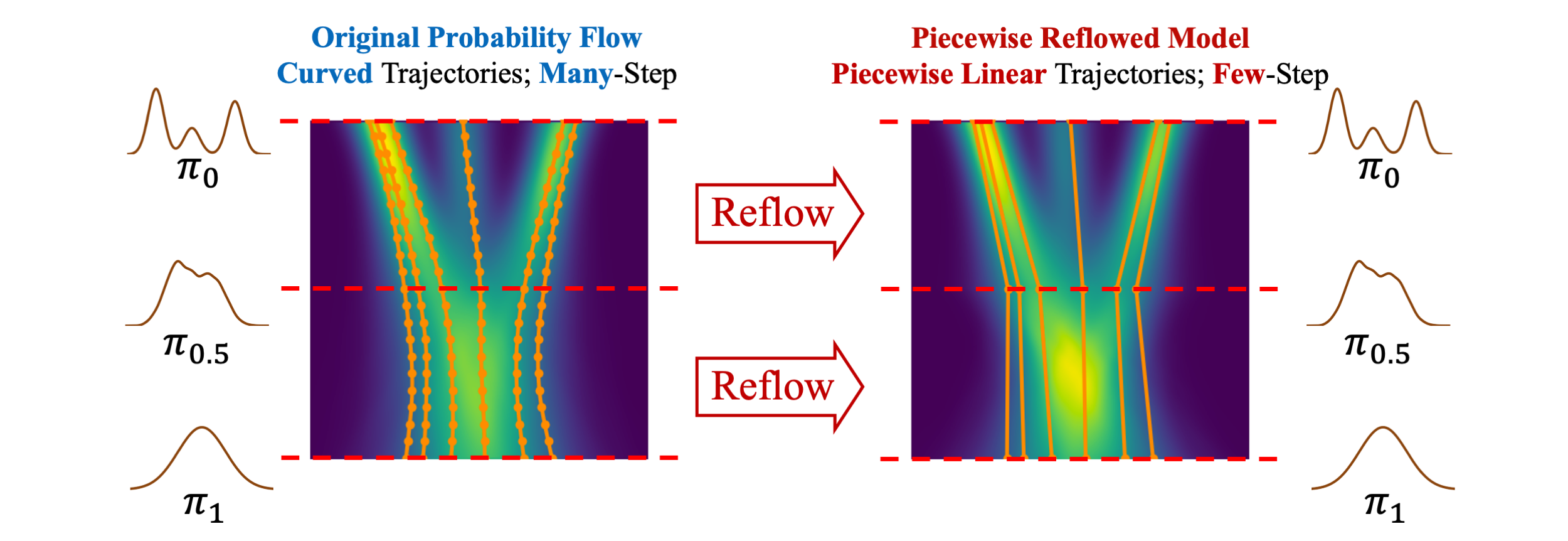}
    \vspace{-5pt}
    \caption{Our few-step generator PeRFlow is trained by a divide-and-conquer strategy. We divide the ODE trajectories into several intervals and perform reflow in each time window to straighten the sampling trajectories.}
    \label{fig:perflow}
    \vspace{-1em}
\end{figure}

We create $K$ time windows $\{[t_{k}, t_{k-1})\}_{k=K}^1$ where $1=t_K >\dots> t_k>t_{k-1}>\dots > t_0=0$. For each time window $[t_k, t_{k-1})$, the starting distribution $\pi_{k}$ will be the marginal distribution of the diffusion process at time $t_k$. It can be derived from $\vz_{t_k} = \sqrt{1-\sigma^2(t_k)} \vz_0 + \sigma(t_k) \epsilon$. The target end distribution $\pi_{k-1}$ is constructed by $\Phi(\vz_{t_k}, t_k, t_{k-1})$. We train the PeRFlow model, denoted by $\theta$, to fit the linear interpolation between $\rvz_{t_k}$ and $\rvz_{t_{k-1}}$ for all $k\in[1,\dots,K]$. 
\begin{equation}
\begin{aligned}
    &~~~~\min_{\theta} \sum_{k=1}^K \E_{\vz_{t_k} \sim \pi_k} \left [ \int_{t_{k-1}}^{t_k} 
    \left \| \frac{\vz_{t_{k-1}} - \vz_{t_k}}{t_{k-1}-t_k} - v_{\theta}(\vz_t, t) \right \|^2 
    \dd t \right ], \\
    \text{with} \quad &\vz_{t_{k-1}} = \Phi(\vz_{t_k}, t_k, t_{k-1}) \quad \text{and} \quad \vz_t = \frac{t - t_{k-1}}{t_{k} - t_{k-1}} \vz_{t_{k}} + \frac{t_{k} - t}{t_{k} - t_{k-1}} \vz_{t_{k-1}}.
    \label{eq:perflow-velmatching}
\end{aligned}
\end{equation}

% $$ z_e = \lambda_s * z_s + \eta_s * \epsilon,  --->  \epsilon = (z_e - \lambda_s * z_s) / \eta_s
% $$

\paragraph{Parameterization} The pretrained diffusion models are usually trained by two parameterization tricks, namely \textit{$\epsilon$-prediction} and \textit{velocity-prediction}. To inherit knowledge from the pretrained network, we parameterize the PeRFlow model as the same type of diffusion and initialize network $\theta$ from the pretrained diffusion model $\phi$. 
For the \textit{velocity-prediction}, we can train the PeRFlow model by velocity-matching in \eqref{eq:perflow-velmatching}. To accommodate \textit{$\epsilon$-prediction}, 
%we redefine the noise $\epsilon(t)$ as \eqref{eq:perflow-eps} for $t \in [t_k, t_{k-1}]$. 
we can represent the denoised state $\vz_{t_{k-1}}$ with the starting state $\vz_{t_k}$ and $\epsilon$:
\begin{equation}
    \vz_{t_{k-1}} = \lambda_{k} \vz_{t_k} + \eta_{k} \epsilon, 
    \label{eq:perflow-eps}
\end{equation}
where $\lambda_{k}>1$ and $\eta_{k}$ are defined by the user. We propose to train a neural network $\epsilon_{\theta}(\vz_t, t)$ to estimate the noise $\epsilon$ in \eqref{eq:perflow-eps} based on $\vz_t$ for all $t\in [t_{k}, t_{k-1})$: 
\begin{equation}    
\begin{aligned}
    &\quad \min_{\theta} \sum_{k=1}^K \E_{\vz_{t_k} \sim \pi_k} \left [ \int_{t_{k-1}}^{t_k} 
    \left \| \frac{\vz_{t_{k-1}} - \lambda_{k} \vz_{t_k}}{\eta_k} - \epsilon_{\theta}(\vz_t, t) \right \|^2 
    \dd t \right ],  \\
    \text{with} &\quad \vz_{t_{k-1}} = \Phi(\vz_{t_k}, t_k, t_{k-1}) \quad \text{and} \quad \vz_t = \frac{t - t_{k-1}}{t_{k} - t_{k-1}} \vz_{t_{k}} + \frac{t_{k} - t}{t_{k} - t_{k-1}} \vz_{t_{k-1}}. 
\end{aligned}
\label{eq:perflow-epsmatching}
\end{equation}

%When sampling, we need the velocity $\vv(\vz_t)$ for solving \eqref{eq:flow-ode}. Thus, we derive the relation between $(\epsilon, \vz_t)$ and velocity $v$ by plugging \eqref{eq:perflow-eps} into \eqref{eq:linear_interpolation}:

The optimum of~\eqref{eq:perflow-velmatching} and~\eqref{eq:perflow-epsmatching} are,
\begin{equation}
    v^*(\vz_t, t) = \E \left[ \frac{\vz_{t_{k-1}}-\vz_{t_k}}{t_{k-1} - t_k} \middle\vert \vz_t \right], \quad \text{and} \quad \epsilon^*(\vz_t, t) = \E \left [ \frac{\vz_{t_{k-1}} - \lambda_k \vz_{t_k}}{\eta_k} \middle \vert \vz_t \right].
    \nonumber
\end{equation}
Using calculus and the fact $\vz_t = \frac{t - t_{k-1}}{t_{k} - t_{k-1}} \vz_{t_{k}} + \frac{t_{k} - t}{t_{k} - t_{k-1}} \vz_{t_{k-1}}$, we get,
\begin{equation}
    \textcolor{red}{v^*}(\vz_t, t) = \frac{(1-\lambda_k)\textcolor{red}{z_t} - \eta_k \textcolor{red}{\epsilon^*}(\vz_t, t)}{t-t_{k-1}+\lambda_k t_k - \lambda_k t}
    \label{eq:perflow-getV}
\end{equation}

% \begin{align}
%     &\vz_{t_{k-1}}  = \lambda_t \vz_t + \eta_t \epsilon \\
%     \text{with} \quad &\lambda_t = \frac{\lambda_k (t_{k-1}-t_k)}{\lambda_k (t_{k-1}-t_k) + (t_{k-1}-t_k)} \\
%     & \eta_t = \frac{\eta_k (t_{k-1}-t)}{(t_{k-1}-t)} \\
%     % \vz_{t} & = ( \frac{t-t_k}{t_{k-1}-t_k} + \frac{1}{\lambda_k} \frac{t_{k-1}-t}{t_{k-1}-t_k} ) \vz_{t_{k-1}}  - \frac{\eta_k}{\lambda_k}\frac{t_{k-1}-t}{t_{k-1}-t_k} \epsilon \\
%     % & \\
%     % \vz_{t} & = \vz_{t_k} + \frac{\vz_{t_k}-\vz_{t_{k-1}}}{t_k - t_{k-1}} (t -t_{k}) \nonumber \\
%     % & = (\lambda_{k-1} + \frac{t-t_{k}}{t_{k-1}-t_k} (1-\lambda_{k-1}))\vz_{t_{k-1}} + (1-\frac{t-t_k}{t_{k-1}-t_k})\eta_{k-1} \textcolor{red}{\epsilon};  \\
%     &\textcolor{red}{v}  = \frac{\vz_{t}-\vz_{t_{k-1}}}{t - t_{k-1}} = \frac{(1-\lambda_t) \textcolor{red}{\vz_t} - \eta_t \textcolor{red}{\epsilon} }{t - t_{k-1}}. \label{eq:perflow-getV}
% \end{align}

The sampling process involves first computing $\epsilon_{\theta}(\vz_{t}, t)$ from $\vz_{t}$, then estimating the velocity $\vv(\vz_t)$ via \eqref{eq:perflow-getV} for solving the ODE \eqref{eq:flow-ode}. In this paper, we consider two choices for $\lambda$ and $\eta$:
\begin{itemize}[leftmargin=10pt, itemindent=5pt]
    \item \textit{Parameterization [A]}: According to the definition of the diffusion process, we have $\vz_{t_k} = \gamma \vz_{t_{k-1}} + \sqrt{1-\gamma^2} \epsilon$ with $\gamma = \sqrt{(1-\sigma^2_k) / (1-\sigma^2_{k-1})}$.
    % Since $\vz_{t_k} = \sqrt{1-\sigma^2_k} \vz_0 + \sigma_k \epsilon$ and  $\vz_{t_{k-1}} = \sqrt{1-\sigma^2_{k-1}} \vz_0 + \sigma_{k-1} \epsilon$, given the same random noise $\epsilon$ and data $\vz_0$, 
    We can represent $\vz_{t_k}$ with $\vz_{t_{k-1}}$ and yield, 
    \begin{equation}
        \lambda_{k}=\frac{\sqrt{1-\sigma^2_{k-1}}}{\sqrt{1-\sigma^2_k}}, \quad 
        % \eta_{k}=-\frac{\sigma_{t_{k}} \sqrt{1-\sigma_{k-1}^2}}{\sqrt{1-\sigma_k^2}} + \sigma_{t_{k-1}}.
        \eta_{k}=-\frac{\sqrt{\sigma_k^2-\sigma_{k-1}^2}}{\sqrt{1-\sigma_{k}^2}}.
        \label{eq:perflow-diff-eps}
    \end{equation}
    
    \item \textit{Parameterization [B]}: We can also follow the DDIM solver~\citep{song_denoising_2022}, i.e., $$\vz_{t_{k-1}} = \sqrt{ \frac{\alpha_{t_{k-1}}}{\alpha_{t_k}}} \vz_{t_k} + \sqrt{\alpha_{t_{k-1}}}\left ( \sqrt{\frac{1 - \alpha_{t_{k-1}}}{\alpha_{t_{k-1}}}} - \sqrt{\frac{1 - \alpha_{t_{k}}}{\alpha_{t_{k}}}} \right ) \epsilon_\theta(\vz_{t_k}, t_k),$$ where $\alpha_k = 1-\sigma^2_k$. We can correspondingly set, 
    \begin{equation}
        \lambda_{k} = \frac{\sqrt{\alpha_{k-1}} }{\sqrt{\alpha_k} },  \quad \eta_{k}=\sqrt{1-\alpha_{t_{k-1}}} - \frac{\sqrt{\alpha_{k-1}} }{\sqrt{\alpha_k} } \sqrt{1-\alpha_k}.
        \label{eq:perflow-ddim-eps}
    \end{equation}
    This parameterization initializes the student flow from the update rule of DDIM, which is equivalent to the Euler discretization of the probability flow ODE. We empirically observe that it gives faster training convergence.
\end{itemize}

\begin{algorithm}[t!]
    \SetKwInOut{Input}{Input}
    \SetKwInOut{Output}{Output}
    \SetKwRepeat{Repeat}{repeat}{until}
    \textbf{Input:}~Training dataset $\mathcal{D}$,
    $\epsilon$- or $\vv$-prediction teacher model $f_\phi$,
    Noise schedule $\sigma(t)$,
    ODE solver $\Phi(z_t, t, s, f_{\phi})$,
    Number of windows $K$,
    student model $\epsilon_{\theta}$ or $\vv_{\theta}$,

    \vspace{.5em}
    Create $K$ time windows $\left \{(t_{k-1}, t_{k}] \right \}_{k=1}^{K}$ with $t_K=1$ and $t_0=0$ \;
    Initialize $\theta = \phi$ \;
    \Repeat{convergence}{
        Sample $z_0 \sim \mathcal{D}$\;
        % Sample $t\sim \mathcal{U}(1,0)$ \;
        Sample $k$ from $\{1,\cdots,K\}$ uniformly, then randomly sample time $t \in (t_{k-1}, t_{k}]$ \;
        Sample random noise $\epsilon \sim \mathcal{N}(\vzero, \mI)$ \;
        Get $\vz_{t_k} = \sqrt{1-\sigma^2(t_k)} \vz_0 + \sigma(t_k) \epsilon$ \;
        Solve the endpoint of the time window $z_{t_{k-1}} = \Phi(\vz_{t_k}, t_k, t_{k-1})$ \;
        Get $\vz_t = \vz_{t_k} + \frac{\vz_{t_k}-\vz_{t_{k-1}}}{t_k - t_{k-1}} (t -t_{k})$ \;
        %Get target noise $\epsilon^* = \frac{z_{t_k} - \gamma z_{t_{k-1}}}{\sqrt{1-\gamma^2}}$ where $\gamma=\frac{\sqrt{1-\sigma^2(t_{k})}}{\sqrt{1-\sigma^2(t_{k-1})}}$ \;
        \eIf{$\epsilon$-prediction}
        {
            Compute loss $\ell = \left \|\epsilon_{\theta}(\vz_t, t) - \frac{\vz_{t_{k-1}} - \lambda_{k} \vz_{t_k}}{\eta_k} \right \|^2$ \;
        }{
            Compute loss $\ell = \left \|\vv_{\theta}(\vz_t, t) - \frac{\vz_{t_k} - \vz_{t_{k-1}}}{t_k - t_{k-1}} \right \|^2$ \;
        }
        Update $\theta$ with gradient-based optimizer using $\nabla_\theta \ell$.
    }
    $\Delta W=\theta-\phi$.
    
    \vspace{.5em}
    \textbf{Return:}~Fast PeRFlow $f_{\theta}$ and $\Delta W$.
    \caption{Piecewise Rectified Flow}
\end{algorithm}

\textbf{Scaling Up with Real Training Data}~~
% InstaFlow accelerates pretrained models by performing reflow on a synthetic dataset. In each training iteration, InstaFlow needs to generate images via ODE solver $\Phi(\vz_1, 1, 0)$ where $\vz_1 \sim \mathcal{N}(\vzero, \mI)$. Synthesizing high-quality images usually takes more than 25 inference steps. The resultant high computational cost makes it unfavorable to train InstaFlow on a large-scale dataset. Besides, the images synthesized by pretrained models have a lower quality than real training data. Because InstaFlow lacks the supervision of real images, it has suboptimal generation quality. 
PeRFlow divides the time range $[1,0]$ into multiple windows. For each window, the starting point $\vz_{t_k}$ is obtained by adding random noise to \textit{real} training data $z_0$, and it only requires several inference steps to solve the ending point $\vz_{t_{k-1}}$. The computational cost is significantly reduced for each training iteration compared to InstaFlow, allowing us to train PeRFlow on large-scale training datasets with fast online simulation of the ODE trajectory. Besides, solving endpoints of a shorter time window $[\vz_{t_k}, \vz_{t_{k-1}})$ has lower numerical errors in comparison to the entire time range. High-quality supervision yields significant improvement in the generation results.

\textbf{Classifier-Free Guidance in Training}~~
Classifier-free guidance (CFG)~\citep{ho2021classifier} is a common technique to improve the generation quality of text-to-image models.
During training, we solve the endpoints $\vz_{t_{k-1}}$ for each time window $[{t_k}, {t_{k-1}})$ in an online manner via an ODE solver $\Phi(\vz_{t_k}, t_k, t_{k-1}, \vc, w)$, where $w\geq 1$ denotes the CFG scale, $\vc$ denotes the text prompt.
CFG is turned off when $w=1$.
PeRFlow supports two modes: \textit{CFG-sync} and \textit{CFG-fixed}:
\begin{itemize}[leftmargin=10pt, itemindent=5pt]
    \item \textit{CFG-sync}: We disable CFG by setting $w=1$ for $\Phi(\vz_{t_k}, t_k, t_{k-1}, \vc, w)$. The obtained PeRFlow model can use similar CFG scales as the pretrained diffusion models to guide the sampling.
    \item \textit{CFG-fixed}: We use a pre-defined $w=w^*>1$ for $\Phi(\vz_{t_k}, t_k, t_{k-1}, \vc, w)$ during training. The obtained PeRFLow model learns to straighten the specific ODE trajectories corresponding to  $\Phi(\vz_{t_k}, t_k, t_{k-1}, \vc, w^*)$. One should use a smaller CFG scale (e.g., 1.0-2.5) to adjust guidance when sampling from PeRFLow trained with \textit{CFG-fixed}.
\end{itemize}
Through empirical comparison, we observe that PeRFlow+\textit{CFG-sync} preserves the sampling diversity of the original diffusion models with occasional failure in generating complex structures, while PeRFlow+\textit{CFG-fixed} trades off sampling diversity in exchange for fewer failure cases.

Our recommendations are as follows:
When using powerful pre-trained diffusion models (e.g., SDXL) and prioritizing generation quality, PeRFLow+\textit{CFG-fixed} is the better choice.
On the other hand, when the goal is to maintain the sampling diversity and adaptability of customized fine-tuned plug-ins, such as Dreamshaper, PeRFLow+\textit{CFG-sync} is the more suitable option.

\textbf{PeRFlow as Universal Plug-and-Play Accelerator}~~
% PeRFlow initializes the student model $\theta$ with the pre-trained diffusion model $\phi$. We perform acceleration by training on the text-to-image task. After convergence, we find that the change of weights $\Delta W = \theta-\phi$ not only accelerates the pretrained models to few-step generators but also supports the few-step inference of other pipelines based on the pretrained diffusion models. For example, PeRFlow-SD-v1.5 $\Delta W$ can accelerate the ControlNet \citep{zhang_adding_2023}, IP-Adaptor \citep{ye2023ip}, and 3D-multiview \citep{long2023wonder3d} pipelines designed for SD-v1.5. The accelerated pipelines achieve nearly lossless generation performance. Refer to section \ref{sec:compatability} for detailed results.
PeRFlow initializes the weights of the student model $\theta$ with the pretrained diffusion model $\phi$. After training with piecewise reflow, we find that the change of weights $\Delta W = \theta - \phi$ can be used to seamlessly accelerate many other workflows pretrained with the diffusion model. For exmaple, $\Delta W$ of PeRFlow+SD-v1.5 can accelerate the ControlNets~\citep{zhang_adding_2023}, IP-Adaptor~\citep{ye2023ip} and multiview generation~\citep{long2023wonder3d} pipelines trained with the original SD v1.5. The accelerated pipelines achieve nearly lossless few-step generation as the original many-step generation. Please refer to Section~\ref{sec:compatability} for detailed results.

%%%%%%%%%%%%%%%%%%%%%%%%%%%%%%%%%%%%%%%%%%%%%%%%%%%%%%%%
% \paragraph{CFG scales for sampling via PeRFlow} 
% \paragraph{On the convergence of acceleration} 
% \paragraph{On the division of time windows} 

\section{Experiments}
We use PeRFlow to accelerate several large-scale text-to-image and text-to-video models, including SD-v1.5, SD-v2.1, SDXL, and AnimateDiff. In this section, we will illustrate the experiment configurations and empirical results.

\textbf{Experiment Configuration}~~
All the PeRFlow models are initialized from their diffusion teachers.
PeRFlow-SD-v1.5 is trained with images in resolution of $512\times 512$ using \textit{$\epsilon$-prediction} defined in \eqref{eq:perflow-diff-eps}. 
PeRFlow-SD-v2.1 is trained with images in resolution of $768 \times 768$ using \textit{$\vv$-prediction}. 
PeRFlow-SDXL is trained with images in resolution of $1024\times 1024$ using \textit{$\epsilon$-prediction} defined in \eqref{eq:perflow-ddim-eps}. 
Images are all sampled from the LAION-Aesthetics-5+ dataset~\citep{schuhmann2022laion} and center-cropped.
We also train PeRFlow-AnimateDiff with video clips in size of $16\times 384\times 384$ using \textit{$\epsilon$-prediction} defined in \eqref{eq:perflow-ddim-eps}.
We randomly drop out the text captions with a low probability ($10\%$) to enable classifier-free guidance during sampling. 
We divide the time range $[0,1]$ into four windows uniformly. For each window, we use the DDIM solver to solve the endpoints with 8 steps. 
We refer to the Hugging Face scripts for training Stable Diffusion \footnote{\scriptsize \url{https://github.com/huggingface/diffusers/tree/main/examples/text_to_image}} to set other hyper-parameters, including learning rate and weight decay. All experiments are conducted with $16$ NVIDIA A100 GPUs.

\subsection{Few-step generation}
\begin{figure}[t!]
    \centering
    \includegraphics[width=1.\linewidth]{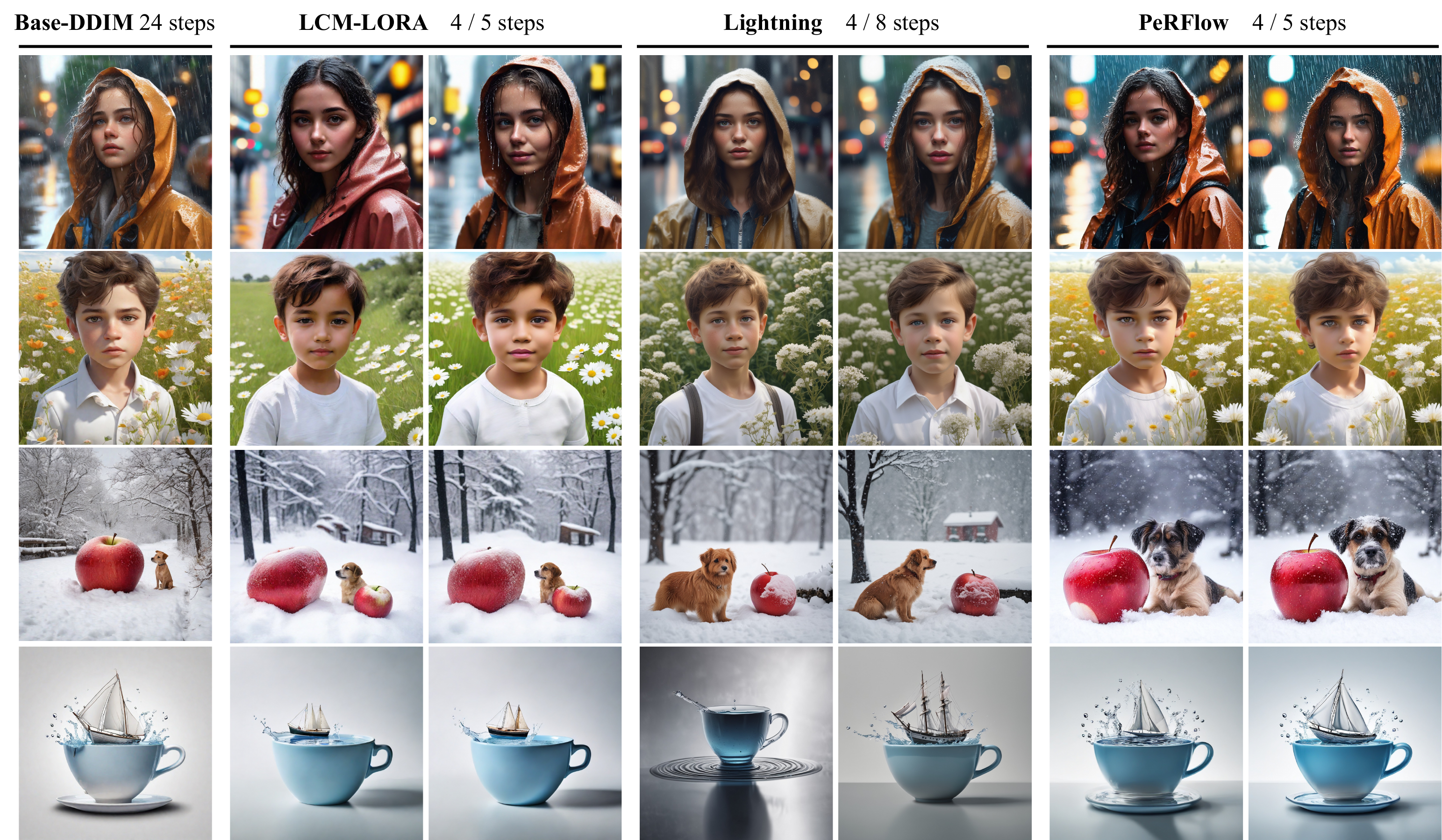}
    \caption{The $1024\times 1024$ images generated by PeRFlow enjoy richer details and better text-image consistency in comparison to other acceleration methods on SDXL. 
    Prompt \#1: \textit{``a closeup face photo of girl, wearing a raincoat, in the street, \underline{heavy rain}, bokeh''}; 
    Prompt \#2: \textit{``a closeup face photo of a boy in white shirt standing on the grassland, flowers''}; 
    Prompt \#3: \textit{``a \underline{huge} red apple in front of a small dog, heavy snow''}.
    Prompt \#4: \textit{``front view of a \underline{boat} sailing in a cup of water''}.
    }
    \label{fig:sdxl-t2i-compare}
    \vspace{-1em}
\end{figure}

\begin{figure}[t!]
    \centering
    \includegraphics[width=0.95\textwidth]{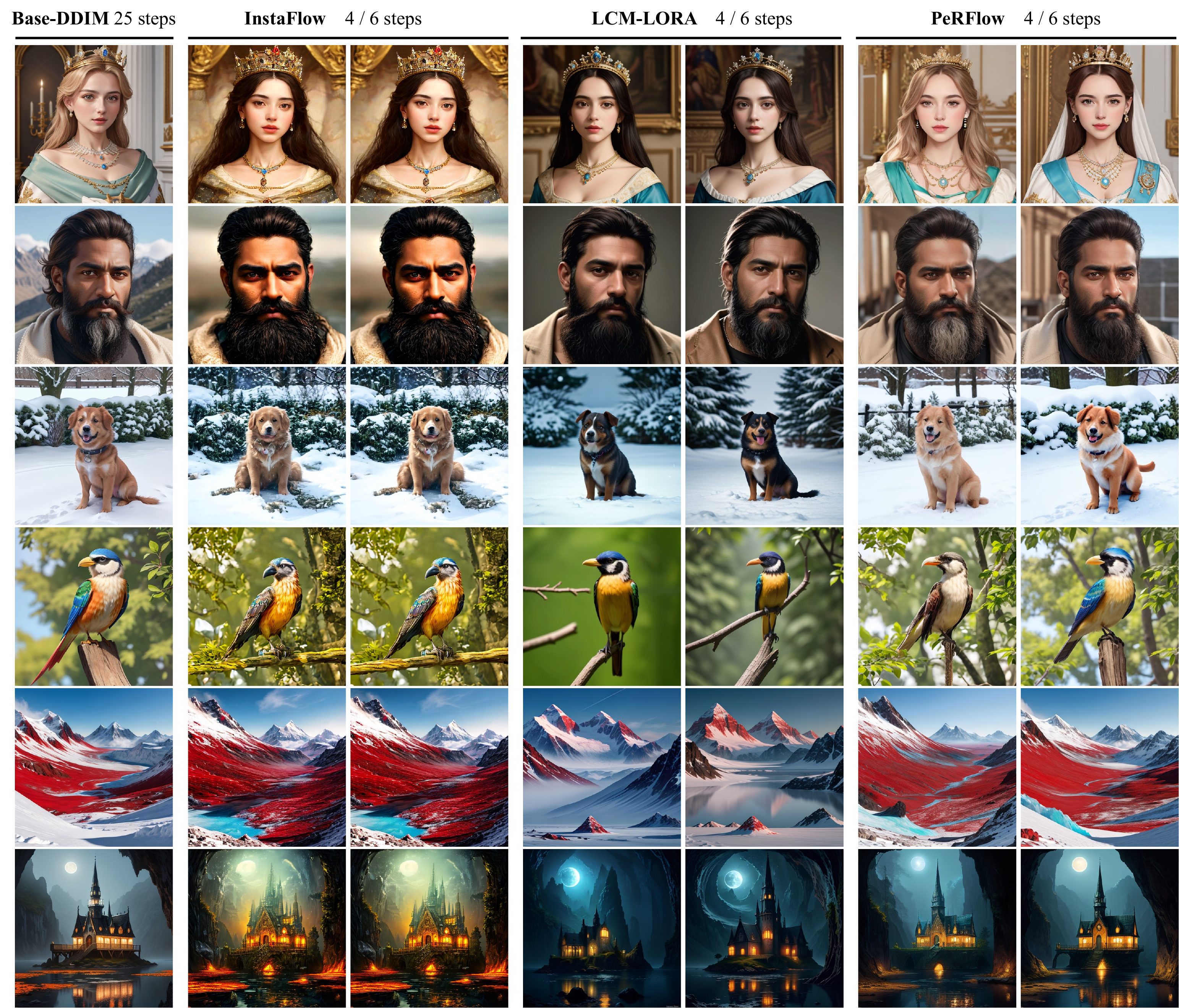}
    \caption{The $512\times 512$ images generated by PeRFlow enjoy richer details and color styles in comparison to other acceleration methods on SD-v1.5 (w/ \href{https://huggingface.co/Lykon}{DreamShaper}). Images in each row are generated with the same random seed.
    }
    \vspace{-0.5em}
    \label{fig:sd15-t2i-compare}
\end{figure}

PeRFlow succeeds in accelerating pretrained Stable Diffusion models to few-step generators. As shown in figure \ref{fig:sdxl-t2i-compare} and \ref{fig:sd15-t2i-compare}, PeRFlow can generate astonishing pictures with only 4 steps. If increasing the number of inference steps (e.g., 5 or 6), we can obtain images with much richer details. We compare the generation results with recent acceleration methods, including InstaFlow, LCM-LORA, and SDXL-lightning. PeRFlow enjoys richer visual texture and better alignment between text prompts and images.
% refer to figure \ref{fig:sd15-t2i-baseline} and \ref{fig:sdxl-t2i-baseline}.
Refer to figure \ref{fig:sd15-t2i-4step}, \ref{fig:sd21-t2i-4step}, and \ref{fig:sdxl-t2i-4step} in \textbf{Appendix} for more results.

We compute the FID values of PeRFlow-accelerated SDs in table \ref{tab:fid} using images on three different reference distributions: (1) LAION-5B-Aesthetics~\citep{schuhmann2022laion}, which is the training set of PeRFlow and other methods; (2) MS COCO 2014~\citep{lin2014microsoft} validation dataset; (3) images generated from SD- v1.5/XL with JourneyDB~\citep{sun2024journeydb} prompts. We generate 30,000 images for the SD-v1.5 models and 10,000 for the SDXL series.
We set the inference steps to 4 and 8 steps, respectively.
In comparison to LCM-LoRA, we observe that PeRFlow models have obviously lower FID values. When increasing the number of inference steps, FID values of PeRFlow decrease because the numerical errors of solving ODE are better controlled. However, FID values of LCM-LoRA unexpectedly increase.

\begin{figure}[t!]
    \centering
    \includegraphics[width=0.93\textwidth]{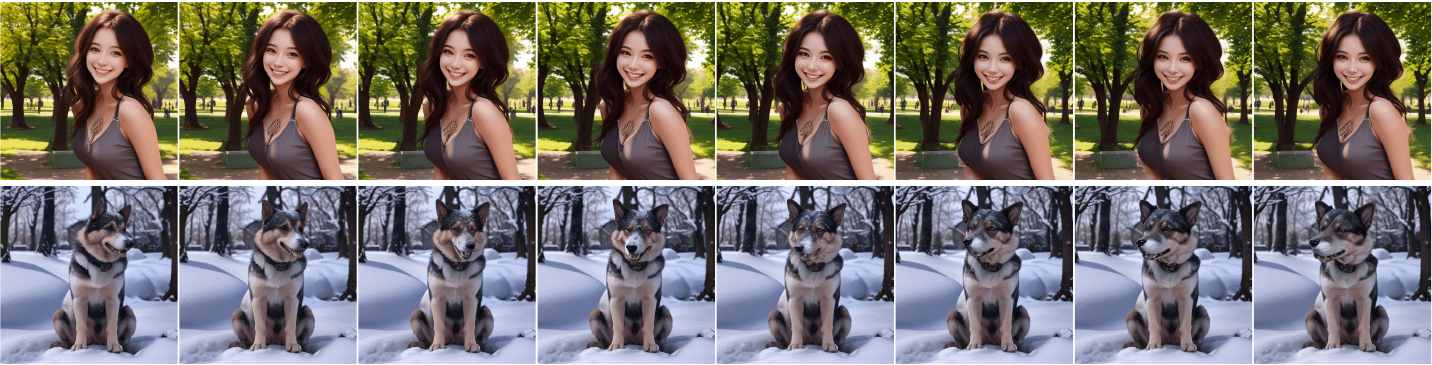}
    \caption{6-step generation ($16\times 512\times 512$) via PeRFlow-AnimateDiff (motion module-v3 with DreamShaper). The text prompts used are \textit{``A young woman smiling, in the park, sunshine''} and \textit{``A dog sitting in the garden, snow, trees''}. 
    %We append \textit{``4k uhd, high quality, masterpiece''} as the prefix of prompts, and use \textit{``distorted, blur, haze, warm, over-saturated, worst quality, low quality, letterboxed''} as the negative prompt.
    }
    \vspace{-1em}
    \label{fig:animatediff-6step}
\end{figure}

\begin{table}[!t]
\centering
\begin{subtable}{0.9\linewidth}
  \centering
  \scalebox{0.9}{
    \begin{tabular}{ccccccc}
        \toprule
           ~  & \multicolumn{2}{c}{LAION-5B} &  \multicolumn{2}{c}{COCO2014} &\multicolumn{2}{c}{SD-v1.5} \\
        \midrule
        \textbf{Method}  & \textbf{4-step}  & \textbf{8-step}  & \textbf{4-step}  & \textbf{8-step}  & \textbf{4-step}  &  \textbf{8-step} \\
        \midrule
        InstaFlow & 14.32 & 10.98 & 13.86 & 11.40 & 16.67 & 10.45 \\
        LCM-LoRA   & 15.28     & 19.21          & 23.49     & 29.63   & 15.63     & 21.19  \\
        PeRFlow    & \textbf{8.60}      & \textbf{8.52}            & \textbf{11.31}     & 14.16  & \textbf{8.28}      & \textbf{5.03}   \\
        \bottomrule
    \end{tabular}
  }
    \caption{\small SD-v1.5}
    % \vspace{-1em}
\end{subtable}

\begin{subtable}{0.9\linewidth}
  \centering
  \scalebox{0.9}{
    \begin{tabular}{ccccccc}
        \toprule
           ~  & \multicolumn{2}{c}{LAION-5B} &  \multicolumn{2}{c}{COCO2014} &\multicolumn{2}{c}{SDXL} \\
        \midrule
        \textbf{Method}  & \textbf{4-step}  & \textbf{8-step}  & \textbf{4-step}  & \textbf{8-step}  & \textbf{4-step}  &  \textbf{8-step} \\
        \midrule
        Lightning & 15.47 & 14.37 & 22.86 & 20.44 & 11.41 & 10.49 \\
        LCM-LoRA   & 13.66 & 13.31 & 19.74 & 21.70 & 9.42 & 9.90 \\
        PeRFlow    & \textbf{13.30} & \textbf{13.06} & \textbf{18.48} & \textbf{19.21} & \textbf{9.28} & \textbf{9.12}  \\
        \bottomrule
    \end{tabular}
  }
  \caption{\small SDXL}
  % \vspace{-1em}
\end{subtable}
\caption{FID values of different acceleration methods.}
\label{tab:fid}
\end{table}

\begin{figure}[th!]
\centering
  \begin{subfigure}{0.32\textwidth}
    \centering
    \includegraphics[width=\linewidth]{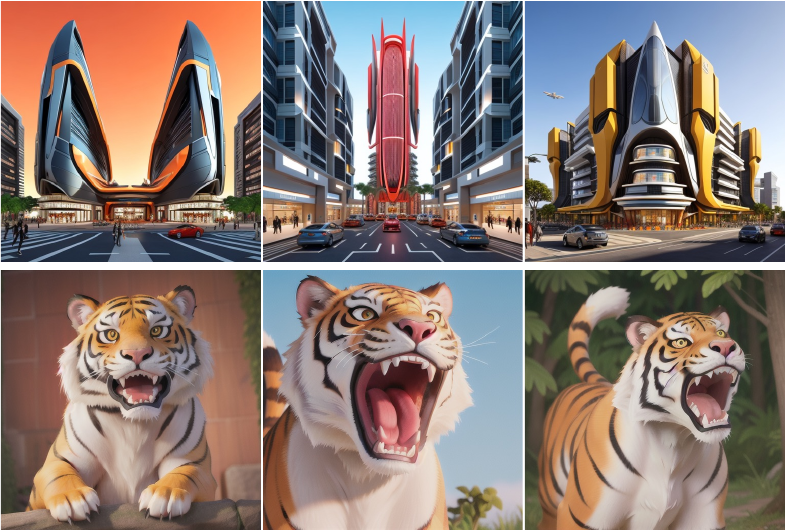}
    \caption{SD-v1.5}
  \end{subfigure} 
  \begin{subfigure}{0.32\textwidth}
    \centering
    \includegraphics[width=\linewidth]{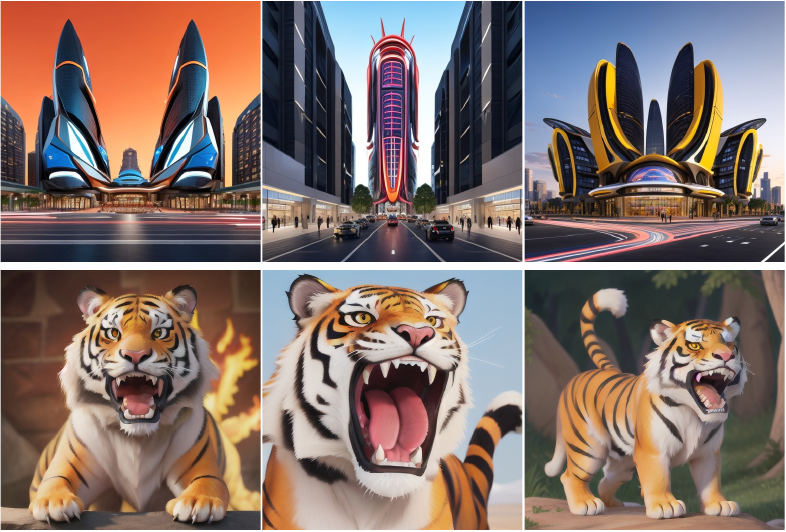}
    \caption{PeRFlow}
  \end{subfigure}
  \begin{subfigure}{0.32\textwidth}
    \centering
    \includegraphics[width=\linewidth]{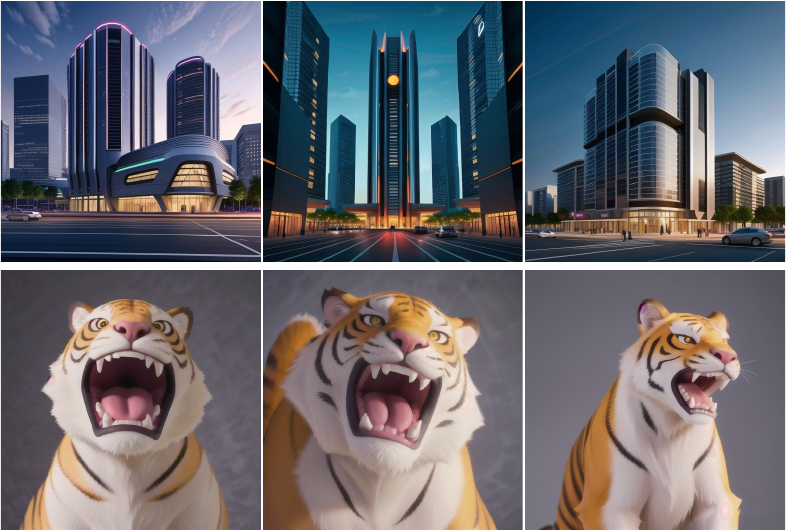}
    \caption{LCM-LoRA}
  \end{subfigure}
  \vspace{-5pt}
  \caption{PeRFlow has better compatibility with customized SD models compared to LCM-LoRA. The top is \href{https://civitai.com/models/114612/architectureexteriorsdlifechiasedamme}{ArchitectureExterior}
  % \footnote{\scriptsize \url{https://civitai.com/models/114612/architectureexteriorsdlifechiasedamme}} 
  and the bottom is \href{https://civitai.com/models/65203/disney-pixar-cartoon-type-a}{DisneyPixarCartoon}.
  % \footnote{\scriptsize\url{https://civitai.com/models/65203/disney-pixar-cartoon-type-a}}
  }
  \label{fig:better_compat}
  % \vspace{-10pt}
\end{figure}

\begin{figure}[th!]
    \centering
  \begin{subfigure}{0.32\textwidth}
    \centering
    \includegraphics[width=\linewidth]{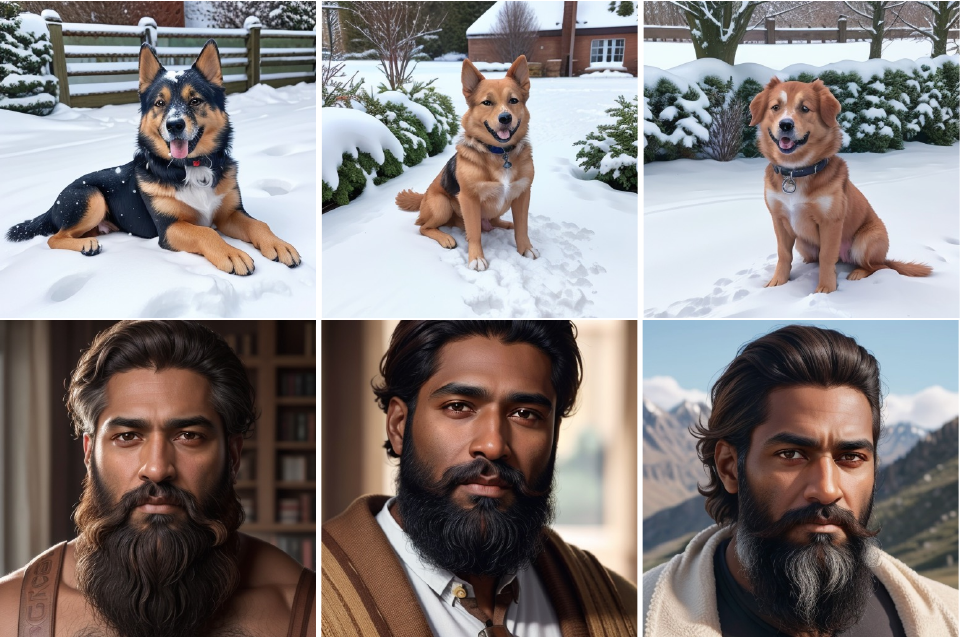}
    \caption{SD-v1.5}
  \end{subfigure} 
  \begin{subfigure}{0.32\textwidth}
    \centering
    \includegraphics[width=\linewidth]{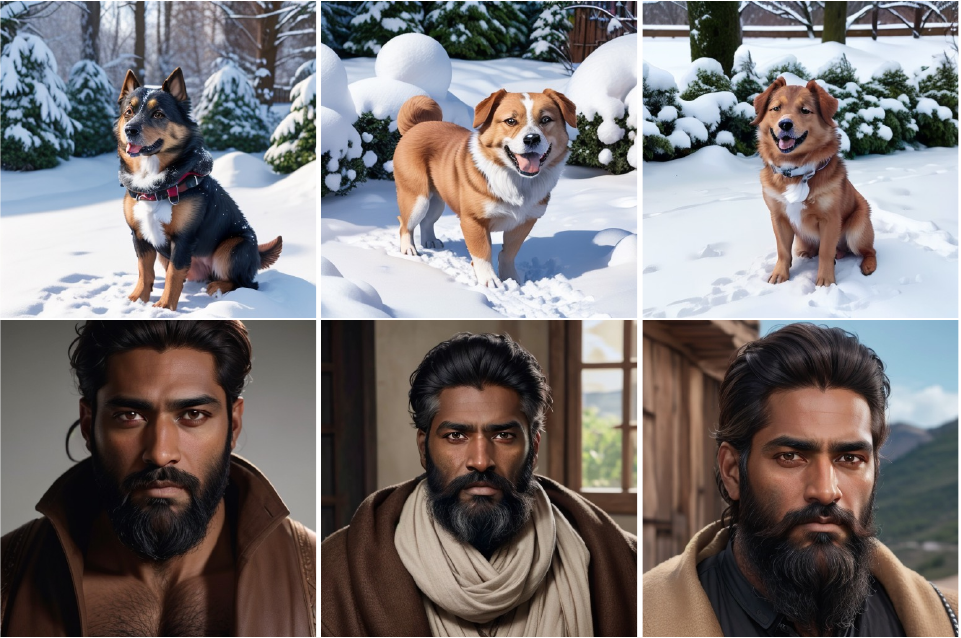}
    \caption{PeRFlow}
  \end{subfigure} 
  \begin{subfigure}{0.32\textwidth}
    \centering
    \includegraphics[width=\linewidth]{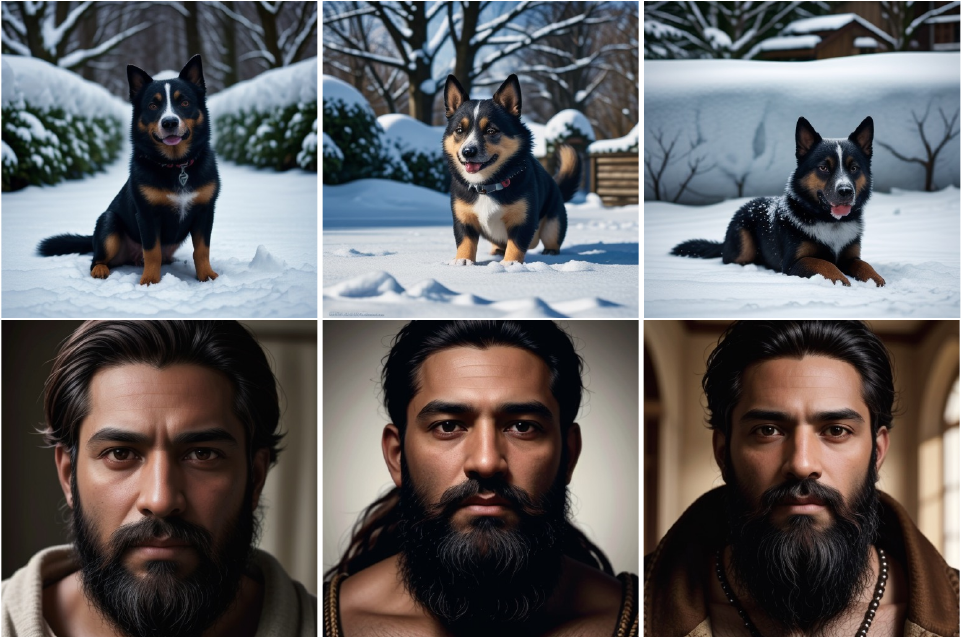}
    \caption{LCM-LORA}
  \end{subfigure}
  \vspace{-5pt}
  \caption{Three random samples from two models with the same prompts. PeRFlow has better sampling diversity compared to LCM-LoRA.}
  \label{fig:better_diversity}
  \vspace{-1em}
\end{figure}

\textbf{Domain shift caused by acceleration}~~
When accelerating diffusion models, we expect to preserve the performance and properties of the pretrained models.
In table~\ref{tab:fid}, we compute the FID values between the generation of the original SD models and the accelerated models. We observe the FID values of PeRFlow are smaller than LCM-LORA, InstaFlow, and SDXL-Lightning. This implies the distribution shift to the original SD models caused by PeRFlow is much smaller than other counterparts. The numerical comparison corresponds to the results in figure \ref{fig:better_compat}. The color style and layout of PeRFlow's results match the results of the pretrained models, while an obvious domain shift appears in the results of LCM-LoRA. Besides, the sampling diversity of PeRFlow is similar to the original SD-v1.5 and appears to be better than LCM-LoRA in figure \ref{fig:better_diversity}.

\subsection{PeRFlow as Universal Plug-and-Play Accelerator on SD Work Flows} \label{sec:compatability}
PeRFlow-$\Delta W$ serves as a universal accelerator that can be simply plugged into various pipelines trained on the pretrained Stable Diffusion models, including (but not limited to) ControlNet \citep{zhang_adding_2023}, IP-Adaptor \citep{ye2023ip}, and multiview generation. For example, plugging PeRFlow-$\Delta W$ into the SD-v1.5 ControlNet-Tile gives a 4-step image enhancement module (figure \ref{fig:sd15-controlnet-tile}). Combining this module with the 4-step PeRFlow-SD-v1.5, we can generate high-quality $1024\times 1024$ images with lightweight SD-v1.5 backbones. 
For multiview generation, plugging PeRFlow-$\Delta W$ into the Wonder3D \citep{long2023wonder3d} pipeline leads to \textbf{one-step} generation of multi-view images (figure \ref{fig:sd15-wonder3d}). More results are shown in figure \ref{fig:sd15-controlnet-others}.
% and \ref{fig:sd15-ip} in the Appendix.

\subsection{Additional Discussion}
\textbf{Inference Budget Allocation}~~
If PeRFlow divides the entire sampling trajectory into $K$ windows and perfectly straightens the sub-trajectories in each window, $K$-step inference (one for each window) will yield high-quality images. However, for pictures with complex structures, such as motorcycles with well-crafted wheels and engines, PeRFlow may require more steps. Ho \etal \citep{ho_denoising_2020} found that diffusion models generate images by synthesizing the layout and structure first and then refining the local details. Inspired by this observation, we allocate the extra steps to windows in highly noisy regions. Supposing we have $K$ time windows $\{[t_{k}, t_{k-1})\}_{k=K}^1$ with $1=t_K >\dots> t_k>t_{k-1}>\dots > t_0=0$, 
we can allocate more than one steps to the earlier windows $\{[t_{k}, t_{k-1})\}_{k=K}^{K_{extra}}$, where $K_{extra}$ is pre-defined.
%Let the number of inference steps be $N <= 2K$,  we allocate $N//K+1$ steps for window $[t_{k}, t_{k-1})$ if $K-k < N\%K$, otherwise $N//K$.
In practice, PeRFlow creates $4$ time windows for acceleration training, and $5$-step inference (with one extra step in $[t_{K}, t_{K-1})$) consistently generates high-quality images. 

\textbf{Dynamic Classifier-Free Guidance}~~
CFG is a useful technique to improve the layout, structure, and text alignment of the generated images. However, a large CFG scale sometimes leads to over-saturated color blocks \citep{kynkaanniemi2024applying, wang2024analysis}. To mitigate this issue, we use a dynamic CFG strategy for few-step sampling, i.e., the corresponding CFG scales decrease for window $K$ to $1$. For example, when sampling with $5$ steps, the CFG schedule is $7.5$-$4.0$-$4.0$-$4.0$ for the \textit{CFG-sync} mode and $2.5$-$1.5$-$1.5$-$1.5$ for the \textit{CFG-fixed} mode.

\section{Related Works}
\textbf{Few-Step Diffusion Models}~~
Diffusion models have demonstrated impressive generative capabilities, but their iterative sampling process often suffers from slow inference speed~\citep{ho_denoising_2020, song_score-based_2021, song_denoising_2022}. To accelerate these models, various methods have been proposed.
Progressive Distillation~\citep{salimans_progressive_2022, meng2023distillation} iteratively reduces the number of inference steps to 4-8, but the error can accumulate during the process.
Alternative approaches~\citep{xiao2021tackling, wang2022diffusion, zheng2022truncated, xu2023ufogen, lin2024sdxllightning, sauer2023adversarial} leverage adversarial losses to align the distributions and reduce the number of inference steps, but these methods often struggle with training instability and mode collapse.
To avoid adversarial training, recent works~\citep{yin2023onestep, nguyen2024swiftbrush, zhou2024score} employ additional models to estimate the score of the generated data for distilling one-step generators, but this adds extra cost to the training pipeline.
Consistency Distillation~\citep{song2023consistency, luo2023latent} is a novel pipeline for distilling few-step diffusion models by optimizing a consistency loss. However, the substantial difference between consistency models and the original diffusion models can hurt their adaptability to pre-trained modules. In our work, PeRFlow provides a simple, clean, and efficient framework for training few-step generative flows. By using different parameterizations as described in Section~\ref{sec:perflow}, PeRFlow achieves minimal gap with diffusion models, making it suitable for various pre-trained workflows.

\textbf{Straight Probability Flows}~~
% Learning straight probability flow is a promising principle to obtain fast generative flows~\citep{liu2022flow, liu2022rectified, liu2023instaflow, finlay2020train}. Reflow is a effective way to learn such straight flows, but it requires constructing a large synthetic dataset~\citep{liu2022flow, liu2023instaflow}, bringing computational overhead and distribution shift. To get rid of dataset construction,~\citep{lee2023minimizing, xing2023exploring} use an extra neural network to estimate the initial noise corresponding to an image, but training this network can be difficult.~\citep{pooladian2023multisample} employs mini-batch optimal transport to directly learn a straighter trajectory, but it is unclear how to apply it to conditional generation scenario such as text-to-image generation.
% \citep{nguyen2023bellman} finds the best stepsize schedule for the pretrained generative model before reflow to improve efficiency, but it cannot avoid dataset generation and the resultant distribution shift.
% PeRFlow gives an new method to avoid using synthetic dataset. It uses real training data to mitigate distribution shift and divide-and-conquer strategy to efficiently perform reflow, leading to advanced few-step text-to-image generator. 
% Sequential reflow~\citep{yoon2024sequential} is a concurrent work to ours. Compared with their work, we additionally provide different parameterization strategies to enhance the empirical performance in accelerating pretrained text-to-image models.
Learning straight probability flow is a promising principle for obtaining fast generative flows~\citep{liu2022flow, liu2022rectified, liu2023instaflow, finlay2020train}. Reflow is an effective way to learn such straight flows, but it requires constructing a large synthetic dataset~\citep{liu2022flow, liu2023instaflow}, which can introduce computational overhead and distribution shift.
To avoid dataset construction,\citep{lee2023minimizing, xing2023exploring} use an extra neural network to estimate the initial noise corresponding to an image, but training this network can be challenging.
\citep{pooladian2023multisample} employs mini-batch optimal transport to directly learn a straighter trajectory, but it is unclear how to apply this method to conditional generation scenarios, such as text-to-image generation.
\citep{nguyen2023bellman} finds the best step-size schedule for the pretrained generative model before reflow to improve efficiency, but it cannot avoid dataset generation and the resulting distribution shift.
PeRFlow provides a new method to avoid using synthetic datasets. It uses real training data to mitigate distribution shift and a divide-and-conquer strategy to efficiently perform reflow, leading to advanced few-step text-to-image generators.
Sequential reflow\citep{yoon2024sequential} is a concurrent work to ours. Compared to their work, we additionally provide different parameterization strategies to enhance the empirical performance in accelerating pre-trained text-to-image models.

\section{Conclusions}
In this work, we present Piecewise Rectified Flow (PeRFlow), a novel technique to learn few-step flow-based generative models. PeRFlow adopts a divide-and-conquer strategy, separating the generation trajectory into intervals and applying the reflow operation within each interval. This yields two key advantages: (1) using real training data to mitigate distribution shift from synthetic data, and (2) avoiding the need to generate and store a synthetic dataset prior to training. PeRFlow also designs proper parameterizations to inherit knowledge from pre-trained diffusion models for fast convergence. Consequently, PeRFlow accelerates powerful diffusion models like SD v1.5, SD v2.1, and SDXL, producing high-quality few-step image generators. Moreover, PeRFlow can be seamlessly combined with various SD workflows to create their accelerated versions.

\textbf{Limitations}~~
Currently, PeRFlow divides the time range into 4 windows, balancing inference and training costs. It needs 4 steps or more for generation. To enable 1-2 step inference, we plan to explore multi-stage training and will focus on avoiding target synthesizing in the future.

\bibliography{references}

%%%%%%%%%%%%%%%%%%%%%%%%%%%%%%%%%%%%%%%%%%%%%%%%%%%%%%%%%%%%
\newpage
\appendix

\begin{center}
    \LARGE \textbf{Appendix}
\end{center}

\section{More generation results}

\begin{figure}[ht!]
    \centering
    \includegraphics[width=1.0\textwidth]{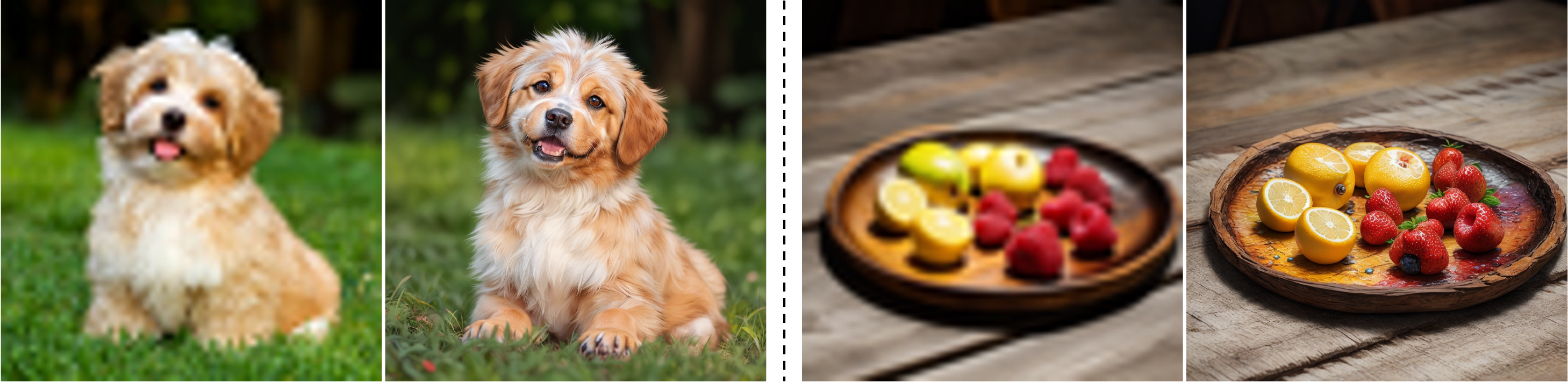}
    \caption{4-step image enhancement ($128 \rightarrow 1024$) with PeRFlow-SD v1.5+ControlNet-tile \citep{zhang_adding_2023}
    }
    \label{fig:sd15-controlnet-tile}
\end{figure}

\begin{figure}[ht!]
    \centering
    \includegraphics[width=1.\textwidth]{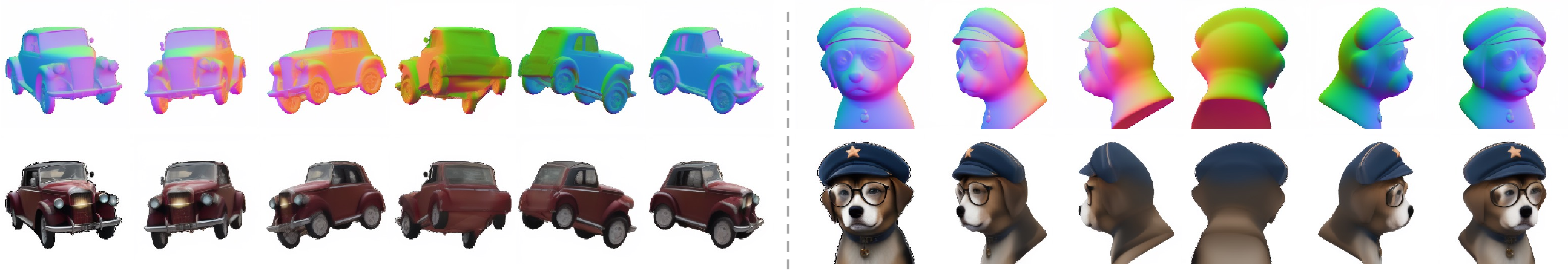}
    \caption{One-step multiview generation of PeRFlow-SD v1.5+Wonder3D~\citep{long2023wonder3d}}
    \label{fig:sd15-wonder3d}
\end{figure}

\begin{figure}[ht!]
    \centering
    \includegraphics[width=0.9\textwidth]{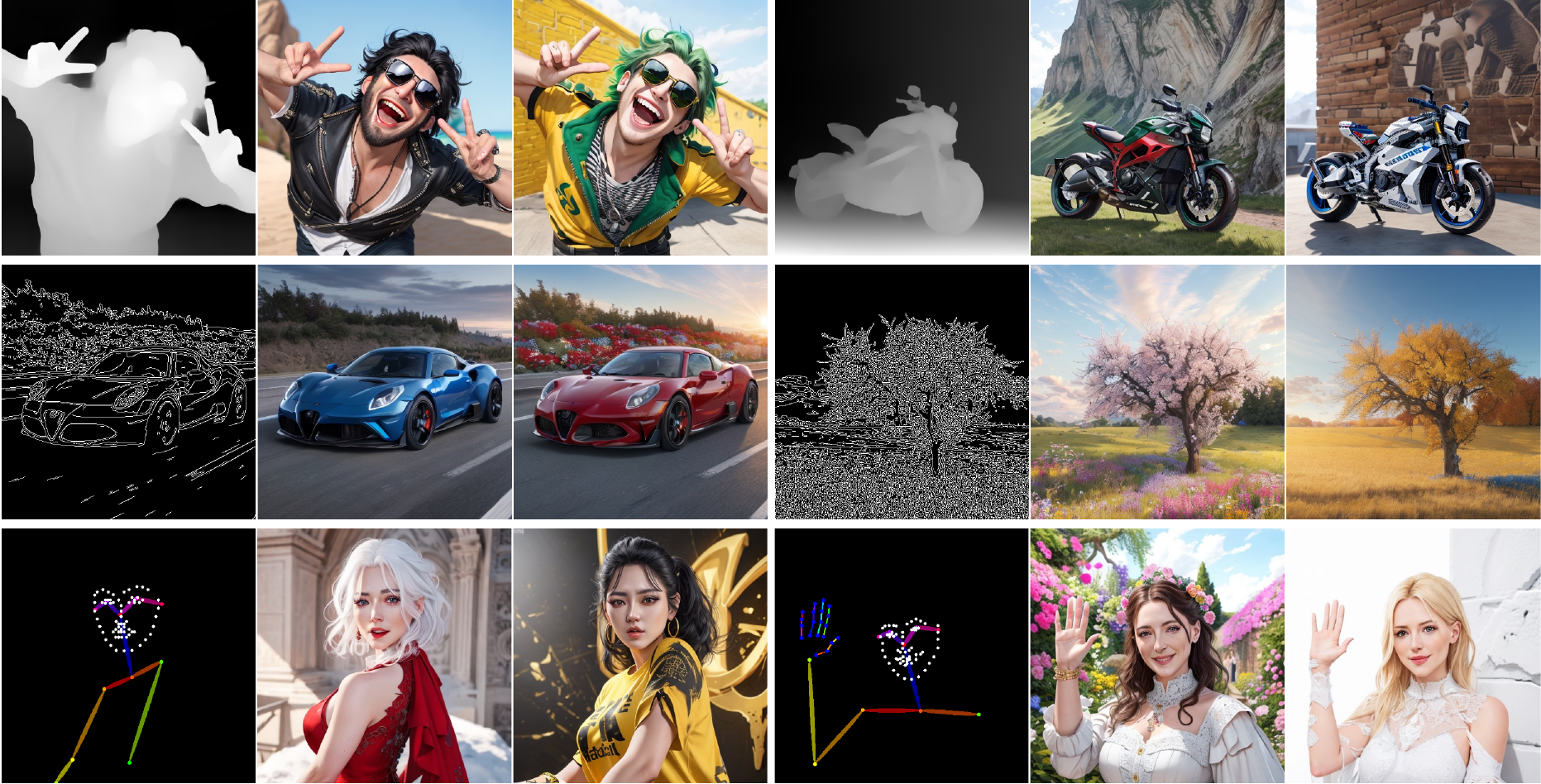}
    \caption{Fast generation via PeRFlow accelerated depth-/edge-/pose-ControlNet \citep{zhang_adding_2023}}
    \label{fig:sd15-controlnet-others}
\end{figure}

\begin{figure}[ht!]
    \centering
    \includegraphics[width=0.95\textwidth]{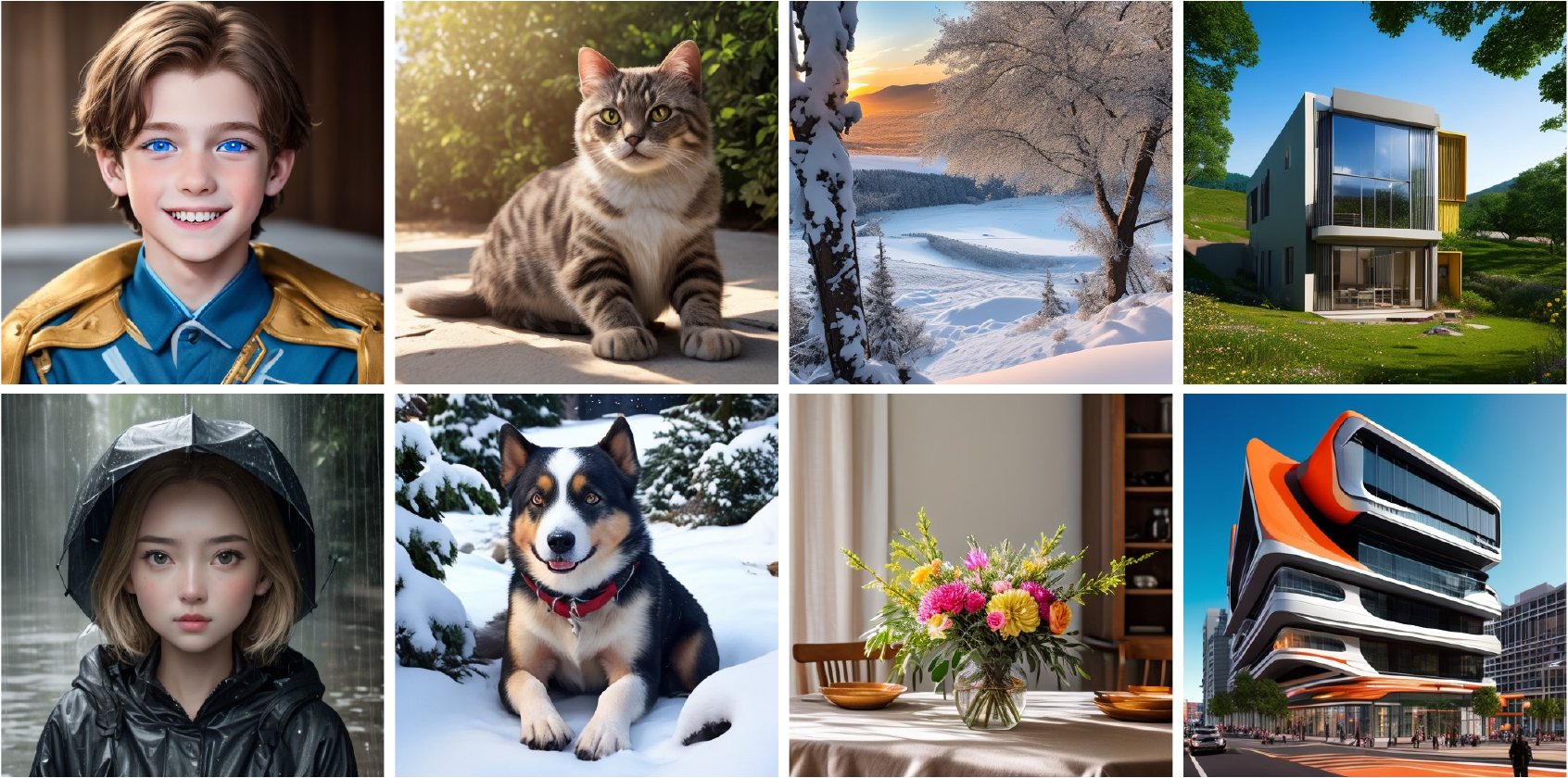}
    \caption{4-step generation ($512\times 512$) via PeRFlow-SD-v1.5.}
    \label{fig:sd15-t2i-4step}
\end{figure}

\begin{figure}[ht!]
    \centering
    \includegraphics[width=0.95\textwidth]{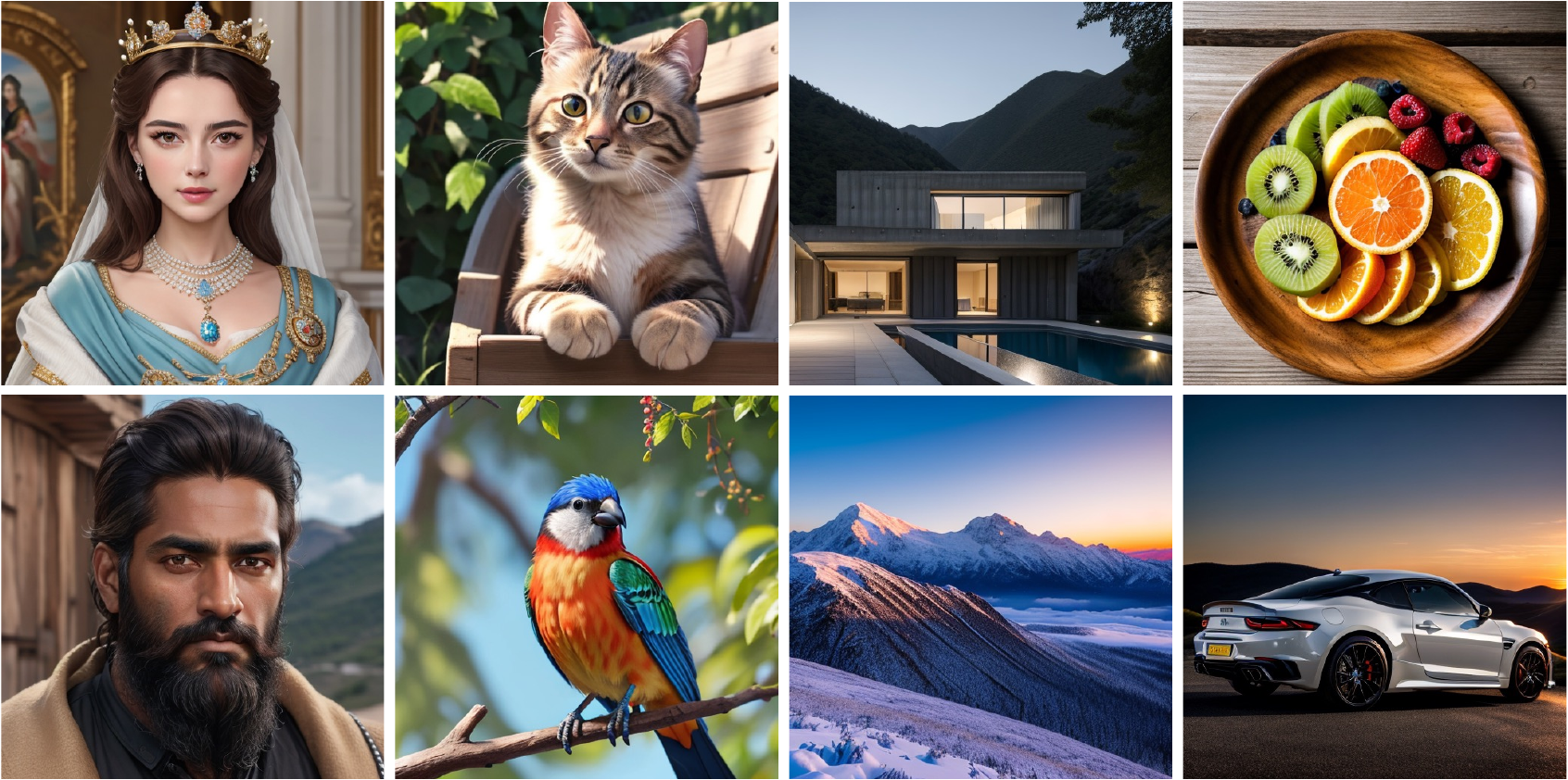}
    \caption{8-step generation ($512\times 512$) via PeRFlow-SD-v1.5.}
    \label{fig:sd15-t2i-8step}
\end{figure}

\begin{figure}[ht!]
    \centering
    \includegraphics[width=0.95\textwidth]{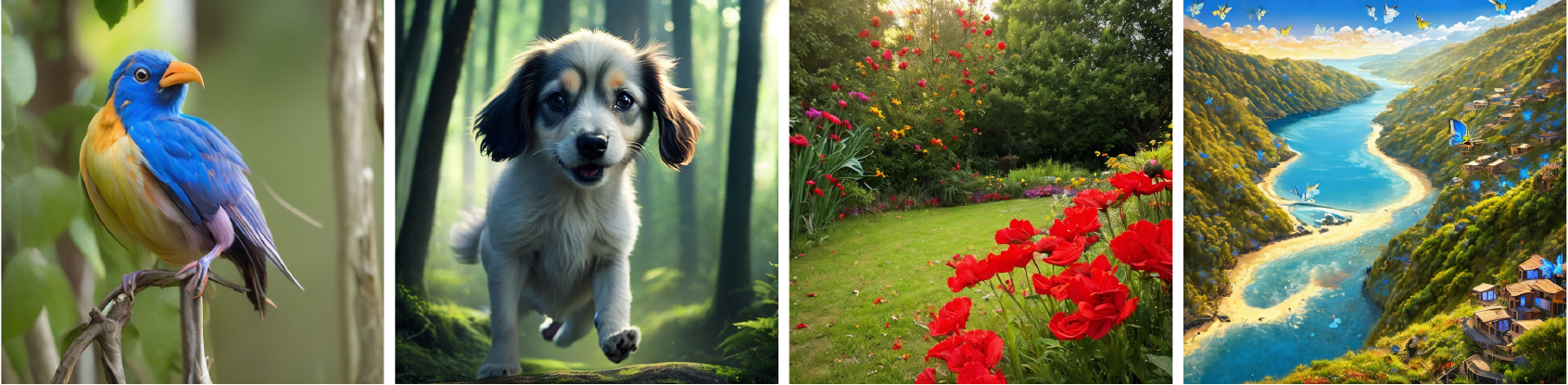}
    \caption{4-step generation ($768\times 768$) via PeRFlow-SD-v2.1.}
    \label{fig:sd21-t2i-4step}
\end{figure}

\begin{figure}[ht!]
    \centering
    \includegraphics[width=0.95\textwidth]{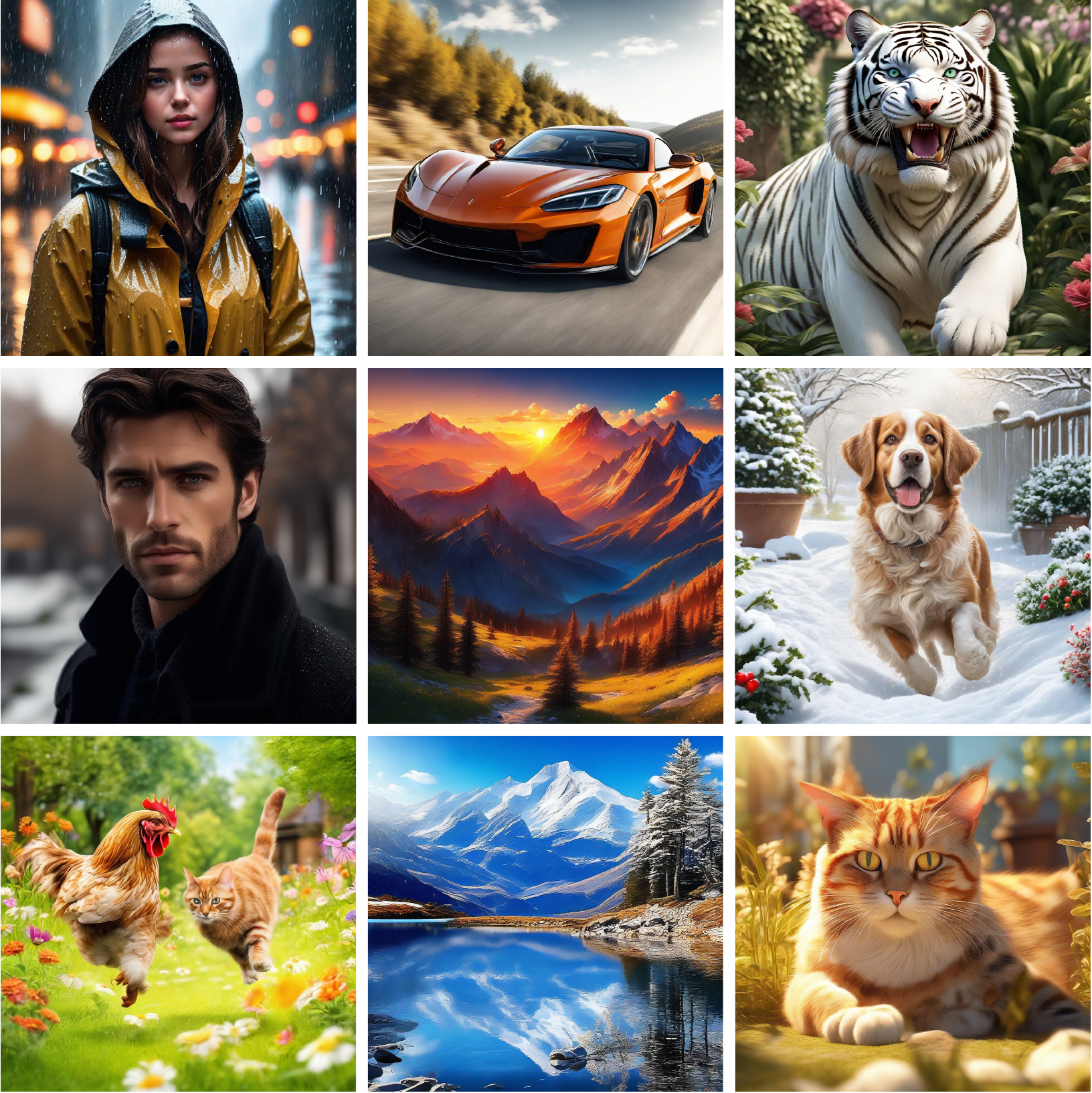}
    \caption{4-step generation ($1024\times 1024$) via PeRFlow-SDXL.}
    \label{fig:sdxl-t2i-4step}
\end{figure}

% \begin{figure}[ht!]
%     \centering
%     \includegraphics[width=0.95\textwidth]{assets/pdf/others/ip_face}
%     \caption{PeRFlow works compatibility with IP-Adaptor.}
%     \label{fig:sd15-ip}
% \end{figure}

\end{document}